\documentclass[twoside]{article}

\usepackage{lipsum}
\usepackage{mathtools}
\usepackage{cuted}
\usepackage{rotating}
\usepackage{longtable}
\usepackage{pdflscape}
\usepackage[accepted]{aistats2022}
\begin{document}

%

%

\twocolumn[

\aistatstitle{Analysis of memory consumption by neural networks based on hyperparameters}

\aistatsauthor{ Mahendran N}

\aistatsaddress{ Indian Institute of Technology, Tirupati} ]

\begin{abstract}
  Deep learning models are trained and deployed in multiple domains. Increasing usage of deep learning models alarms the usage of memory consumed while computation by deep learning models. Existing approaches for reducing memory consumption like model compression, hardware changes are specific. We propose a generic analysis of memory consumption while training deep learning models in comparison with hyperparameters used for training. Hyperparameters which includes the learning rate, batchsize, number of hidden layers and depth of layers decide the model performance, accuracy of the model. We assume the optimizers and type of hidden layers as a known values. The change in hyperparamaters and the number of hidden layers are the variables considered in this proposed approach. For better understanding of the computation cost, this proposed analysis studies the change in memory consumption with respect to hyperparameters as main focus. This results in general analysis of memory consumption changes during training when set of hyperparameters are altered.
\end{abstract}

\section{INTRODUCTION}
Deep learning has achieved impressive results in computer vision \cite{Krizhevsky12,Kaiming16}, healthcare applications \cite{esteva19}, autonomous vehicles \cite{grigorescu20} and so on. The increase in adaptation of deep learning models is indirectly increasing the computation power needed for devices to perform training of deep learning models. Even after providing those requirements, the time taken for training the models is high, leading to increased consumption of energy and memory \cite{garcia2019estimation}. Implementing deep learning algorithms in ultra-low precision 4-bit training opens a new area of research in developing solutions using deep learning \cite{Sun20}. However, the memory consumption has to be taken care of when training is also implemented in less computational environments.

In order to enable neural network training in the hardware limited devices reducing the memory consumption is one of the ways to improvise the model. The existing research on reducing memory consumption have been tried on data centers \cite{Dayarathna15}, mobile applications \cite{Kim15}. In deep learning, analysing memory consumption while training the models is a novel approach taken towards reducing the memory consumption. Related researches exist on trained models in analysing power consumption of deployed models, reducing memory size and retaining accuracy when model is compressed \cite{Kim15}. The models are trained on the server environment and then implemented on less computation devices. In order to train the models itself in less computational environment, we need to analyse memory consumption in existing approaches. This initial step can give better insights of memory consumption by models. We propose an novel approach of analysing memory consumption of deep learning models by considering hyperparameters as variables.

The memory consumption by varying hyperparameters is analysed on three datasets namely MNIST, Fashion MNIST and CIFAR-10. We use Dense layer, Convolutional neural network layer for this analysis. The average memory consumption is calculated by varying the major hyperparameteres like learning rate, number of iterations(epoch), batch size, number of hidden layers and nodes in the hidden layer. The rest of hyperparameters like kernel size, stride are considered as constant. When training the neural network with the above variables, the memory consumption data is fetched with an interval of 5 seconds. The average of memory consumption data is taken for each trail. The results are analysed in comparison with the variables within dataset and outside of the dataset.

\section{LITERATURE REVIEW}
isting approaches taken in field of memory consumption focuses on deployed models or while training. For deployed models, the power consumption, model compression techniques are applied for reducing model size. For training the models, the memory consumption are analysed in order to reduce it. There are both hardware and software approaches taken in each models.
\subsection{Model Compression}
Memory consumption approach is not needed for deployed models but the deployed models consumes space and power. Existing researches on deployed models focuses on increasing speed, model compression techniques. For increasing the speed of neural network on CPUs, researches follow a specific approaches like batch computation to increase the speed \cite{Vanhoucke11}. Compression techniques are implemented in order to reduce the size of deep learning models. As the pre trained model occupies lot of space, compression techniques can compress the models with negligible loss in accuracy from research \cite{Louizos17}. \cite{Kim15} analyses models trained on Titan X and smartphone can results in model compression for significant reduction in model size, runtime and energy consumption. \cite{Chen16} reduces the memory consumption in deep learning by using algorithmic approach in training Deep nets.

All these compression techniques on deep learning models are done on the fully trained, ready-to-deploy models. New advancements like training entire deep learning algorithms in 4-bit processor paves the way that in future models can be trained in least computational devices. This is the major reason for analysing the memory consumption by deep learning algorithms

\subsection{Memory Consumption}

\cite{Pleiss17} research follows memory sharing operations which reduces memory consumption in training. This approach paves the way for reducing memory consumption while training. Analysis of memory consumption in training will give better understanding of the link between the memory consumption and neural network models training. In hardware approach to solve this problem, there have been prototype chip for memory efficient deep learning \cite{Liu18}. \cite{Gao20} research on estimation of GPU memory consumption for deep learning proposes an estimation tool. This tool helps in calculating the memory consumption of both the computation graph and the Deep Learning framework runtime. Instead of improvising a existing deep learning approach, reserchers speed up the deep learning training which in turn reduces the memory consumption. \cite{Liu19} developed point-voxel cnn, a 3D deep learning. They claim to have 1.5X speed up of training which directly reduces GPU memory reduction. Researches are proposing algorithmic approach to make memory efficient models \cite{Tokui19,Zhang19}. There are memory saving architectures which claims that they can reduce the total energy consumption by $26.0\%$ by reducing the refresh energy consumption while maintaining accuracy and high performance \cite{Nguyen20}.

\section{METHOD}
This section introduces the details of the memory consumption extraction and list of configured deep learning models choosen for analysis are discussed.

\subsection{Memory consumption}
Inspired from the power consumption tools / softwares for reducing power consumption in mobile applications and data centers, the idea of memory consumption in terms of deep learning algorithms is explored. In general, the memory consumption in calculated with the formula \ref{eq:gen},

This same method is used here for calculating memory consumption. Deep learning model consumes memory when the model is trained. The consumption varies according to the configuration of neural network. Memory consumption has to run simultaneously in order to provide the data. The consumption data needed has to be extracted in real-time when model is still training in the same system. Two process should run without hassle and provide extracted data.

Multiprocessing is the only possible way to give real time data for visualization. In python, package named ’multiprocessing’ helps in executing process simultaneously. This package supports spawning processes. Spawning function loads and executes a new child process. The current process and new child process can continue to execute concurrent computing. Here multiprocessing in python uses API similar to the threading module. The ’multiprocessing’ package offers concurrency on both locally and remote. We use multiprocessing package in python for this project. One process takes care of extracting memory consumption and another process will be used for training the deep learning model.

Memory consumption is basically memory read/write. These values are extracted from the system. We utilize "psutil" package in python for extracting memory used data. This package gives total memory, memory available, memory percentage, memory used and memory active values. The memory extraction process is done with an interval of 5 seconds. Then the average memory consumption is calculated with the extracted memory used data with the use of formula \ref{eq:avg},

\begin{strip}
\begin{equation}
    \centering
    Memory used = Total memory - (Free memory + Buffers + Cached memory)
    \label{eq:gen}
\end{equation}
\begin{equation}
    Average\ memory\ consumption = \frac{Sum\ of\ memory\ used\ for\ each\ trail}{Number\ of\ trails}
    \label{eq:avg}
\end{equation}
\end{strip}

\subsection{Deep learning models}

Deep learning models are trained when memory consumption process is working simulataneously. Hyperparameters are responsible in developing better deep learning models. We consider changing the hyperparameter values to analyze the memory consumption. Random values are choosen for the list of hyperparamaters learning rate, batch size, number of
epochs and number of layers and number of nodes in each hidden layer. We consider experimenting on three different datasets namely MNIST, Fashion MNIST and CIFAR-10. We use multilayer perceptron networks for training.

The values are choosen close to natural values. i.e., learning rate has to be <0.1 since it is important that neural network learns at each step and if it is too large it can cause the model to converge too quickly at sub optimal solution. Thus we choose the values accordingly. We experiment 5 trails on each dataset and readings are recorded.

\section{EXPERIMENTATION}
The setup, experimentation results, the analysis of memory consumption are discussed in this section.

\textbf{Experimental setup} We use 16GB RAM for training the neural network and analysing the memory consumption. The same setup is utilized throughout for all the datasets.

We experimented on three different datasets namely mnist, fashion mnist and cifar10. Random values are taken for the list of paramaters learning rate, batch size, number of epochs and number of layers. We use memory consumption average value for each trail to train the model. The memory consumption average is different for each model. In general, memory used is considered and will be retrieved. The average is compared with other trails having different parameters. 

\subsection{MNIST dataset}
For MNIST handwritten dataset, the values that change for each trail are learning rate, batch size, number of epochs and number of layer. The hyperparameter values and the memory consumption are tabulated in Table \ref{tab:mnisthyp}

\begin{landscape}
\begin{table}
\centering
\begin{tabular}{llllllllllllll}
 \hline \\
 S.No & learning rate & epochs & batch size & HL 1 & HL 2 & HL 3 & HL 4 & HL 5 & HL 6 & HL 7 & HL 8 & HL 9 & Memory used\\ [3ex]
 \hline \\
    1 & 0.7886 & 18 & 8 & 285 & 389 & 423 & 173 & 61 & - & - & - & - & 2516.309735\\[1.5ex]
    2 & 0.0583 & 25 & 40 & 48 & 236 & 233 & 318 & 245 & 139 & - & - & - & 2519.748392\\[1.5ex]
    3 & 0.0005 & 21 & 97 & 141 & 191 & 150 & 84 & 125 & 148 & 39 & - & - & 2506.735677\\[1.5ex]
    4 & 0.0009 & 46 & 98 & 149 & 310 & 180 & 265 & 163 & - & - & - & - & 2585.272978\\[1.5ex]
    5 & 0.0042 & 22 & 66 & 80 & 60 & 168 & 175 & 54 & 115 & 166 & 321 & 231 & 2533.35625\\[1.5ex]
    6 & 0.00042 & 22 & 66 & 80 & 60 & 168 & 175 & 54 & 115 & 166 & 321 & 231 & 3134.870028\\[1.5ex]
    7 & 0.00087 & 35 & 90 & 252 & 196 & 244 & 215 & 163 & 339 & - & - & - & 2854.54834\\[1.5ex]
    8 & 0.0006 & 24 & 50 & 178 & 12 & 444 & 8 & 200 & 318 & 58 & - & - & 2931.835938\\[1.5ex]
    9 & 0.00012 & 26 & 47 & 248 & 27 & 203 & 398 & 281  & - & - & - & - & 2992.872013\\[1.5ex]
    10 & 0.00065 & 28 & 88 & 150 & 5 & 157 & 216 & 398  & - & - & - & - & 3011.547743\\[1.5ex]
    11 & 0.0012 & 15 & 27 & 138 & 37 & 207 & 326 & 232  & - & - & - & - & 3064.187174\\[1.5ex]
 \hline
\end{tabular} \\[1.5ex]
\caption{Hyperparameter values choosen for MNIST dataset to calculate memory consumption. HL (N) represents number of nodes in Nth hidden layer.}
\label{tab:mnisthyp}
\vspace{1.5em}
\end{table}
\end{landscape}

From the table results of memory consumed, the least memory is consumed by the network which has nodes <200 in each layer. The layer depth, epochs, batch size are varible from the trails and do not contribute to the memory consumption. In terms of learning rate, smaller learning rate leads to high memory consumption.

\begin{figure}[h]
    \centering
    \includegraphics[width=\linewidth,height=12cm]{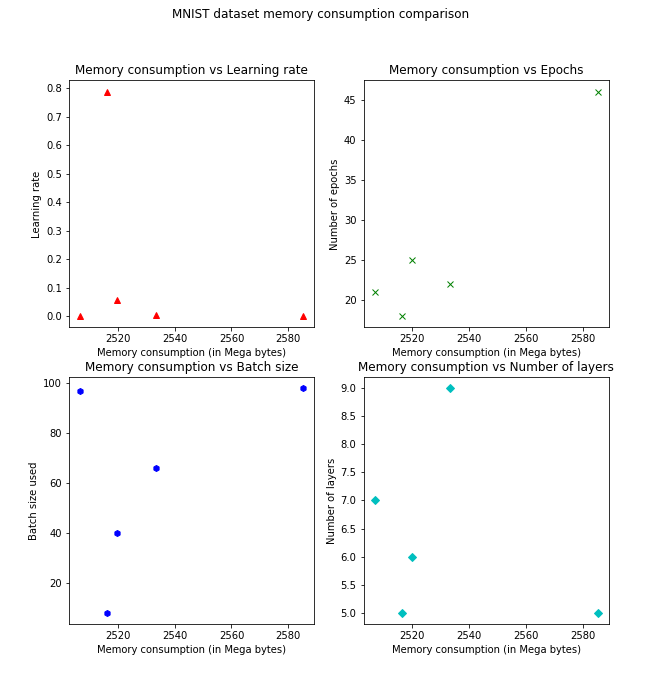}
    \caption{Graph displaying comparison of memory consumed by neural network with the hyperparameter values.}
    \label{fig:mnistres}
\end{figure}

With the considered values, MNIST dataset is trained for memory consumption. The average of memory consumed data is calculated. Results are compared with the hyperparameter values for better understanding of memory consumed. Plotting the five values as graph in figure \ref{fig:mnistres} represents the comparison between the parameters changed and memory consumed by neural network.

\subsection{Fashion MNIST dataset}

For the Fashion MNIST dataset, the values choosen as hyperparameters and the resultant memory consumption are tabulated in Table \ref{tab:fmnisthyp}. Similarly, the number of nodes for each hidden layer is chosen randomly. The values are tabulated in Table  \ref{tab:fmnisthl}. From the tabulated values, the neural network is trained in same environment.

\begin{table}[h]
\begin{tabular}{| c | c | c | c | c |} 
 \hline
 S.No & learning rate & epochs & batch size & Memory consumption \\ [1ex]
 \hline
    1 & 0.3617 & 16 & 17 & 2655.838252 \\[0.5ex]
    2 & 0.1482 & 6 & 11 & 2701.582899 \\[0.5ex]
    3 & 0.0004 & 22 & 6 & 2660.471612 \\[0.5ex]
    4 & 0.0002 & 30 & 18 & 2693.458608 \\[0.5ex]
    5 & 0.0016 & 12 & 11 & 2611.620271 \\[0.5ex]
 \hline
\end{tabular}
\caption{Random hyperparameter values choosen for Fashion MNIST dataset}
\label{tab:fmnisthyp}
\end{table}

\begin{figure}[h]
    \centering
    \includegraphics[width=\linewidth,height=12cm]{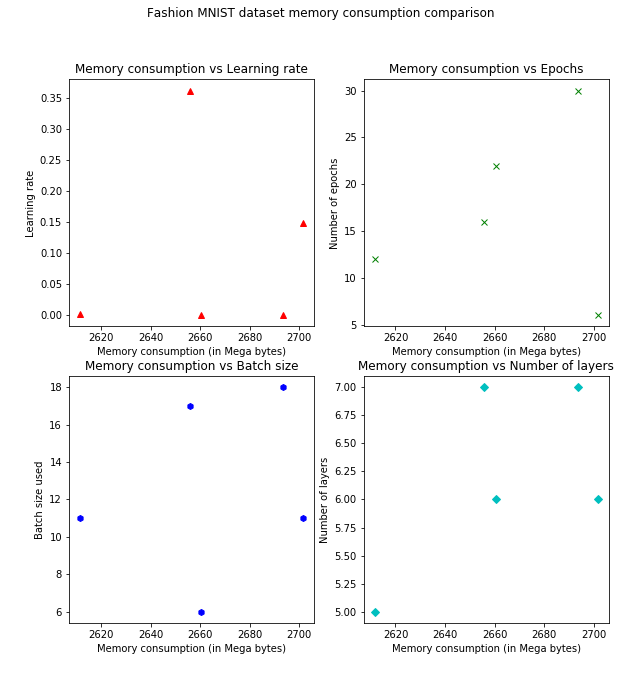}
    \caption{Graph displaying comparison of memory consumed by neural network with the hyperparameter values for Fashion MNIST dataset.}
    \label{fig:fmnistres}
\end{figure}

Using these hyperparameter values, the neural network is trained for the number of iterations as mentioned. The memory utilization value is calculated as mentioned in the previous section. The resultant graph comparing the memory consumption with the hyperparameters chosen is shown in Figure \ref{fig:fmnistres}.

\begin{table}[!h]
\centering
\begin{tabular}{| c | c | c | c | c | c | c | c | c | c |} 
 \hline
 S.No & Layer 1 & Layer 2 & Layer 3 & Layer 4 & Layer 5 & Layer 6 & Layer 7 & Layer 8 & Layer 9 \\ [1ex]
 \hline
1 & 439 & 176 & 388 & 307 & 326 & 160 & 482 & - & -\\[0.5ex]
2 & 349 & 177 & 465 & 506 & 52 & 107 & - & - &- \\[0.5ex]
3 & 181 & 107 & 137 & 259 & 334 & 243 & - & - &- \\[0.5ex]
4 & 362 & 316 & 347 & 380 & 245 & 124 & 195 & - &- \\[0.5ex]
5 & 218 & 45 & 108 & 400 & 121 & - & - & - & -\\[0.5ex]
 \hline
\end{tabular}
\caption{Value of number of nodes choosen in hidden layer for training Fashion MNIST dataset}
\label{tab:fmnisthl}
\end{table}

The resultant graph represents the memory consumption data for the randomly chosen hyperparameter inputs. The resultant data is comparable to the MNIST memory consumption. Here the least memory consumed is 2611 MBs when compared to 2506 MBs in MNIST. From the results, the learning rate do not have much impact the neural networks. But from the values in rows 2 and 4, the increased number of nodes in each hidden layer directly increases the memory consumption. Thus it is evident that if the neural network size is increased horizontally, the memory consumption is more as it contributes to high dimensional multiplication at each iteration.

\subsection{CIFAR 10 dataset}
The content of CIFAR-10 are discussed earlier. With the dataset, the values of hyperparameters and number of layers are tabulated in the Table \ref{tab:cifar10hyp} and Table \ref{tab:cifar10hl}.

\begin{table}[h]
\centering
\begin{tabular}{| c | c | c | c | c |} 
 \hline
 S.No & learning rate & epochs & batch size & Memory consumption \\ [1ex]
 \hline
    1 & 0.0014 & 16 & 14 & 4318.107143 \\[0.5ex]
    2 & 0.0019 & 11 & 7 & 3963.642543 \\[0.5ex]
    3 & 0.0069 & 12 & 13 & 3950.009931 \\[0.5ex]
    4 & 0.0005 & 7 & 26 & 3830.008464 \\[0.5ex]
    5 & 0.0062 & 13 & 8 & 3982.532723 \\[0.5ex]
 \hline
\end{tabular}
\caption{Hyperparameter values choosen for CIFAR-10 dataset}
\label{tab:cifar10hyp}
\end{table}

\begin{table}[h]
\centering
\begin{tabular}{| c | c | c | c | c | c | c | c | c | c |} 
 \hline
 S.No & Layer 1 & Layer 2 & Layer 3 & Layer 4 & Layer 5 & Layer 6 & Layer 7 & Layer 8 & Layer 9 \\ [1ex]
 \hline
1 & 233 & 425 & 299 & 363 & 102 & & & & \\[0.5ex]
2 & 124 & 150 & 273 & 191 & 301 & 293 & 92 & 483 & 65 \\[0.5ex]
3 & 318 & 312 & 412 & 254 & 387 & 60 & 183 & 62 & 132 \\[0.5ex]
4 & 256 & 128 & 256 & 128 & 128	 &  &  &  & \\[0.5ex]
5 & 484 & 195 & 394 & 163 & 322 & 126 & 165	 &  & \\[0.5ex]
 \hline
\end{tabular}
\caption{Value of number of nodes choosen in hidden layer for training CIFAR-10 dataset}
\label{tab:cifar10hl}
\end{table}

\begin{figure}[h]
    \centering
    \includegraphics[width=\linewidth,height=12cm]{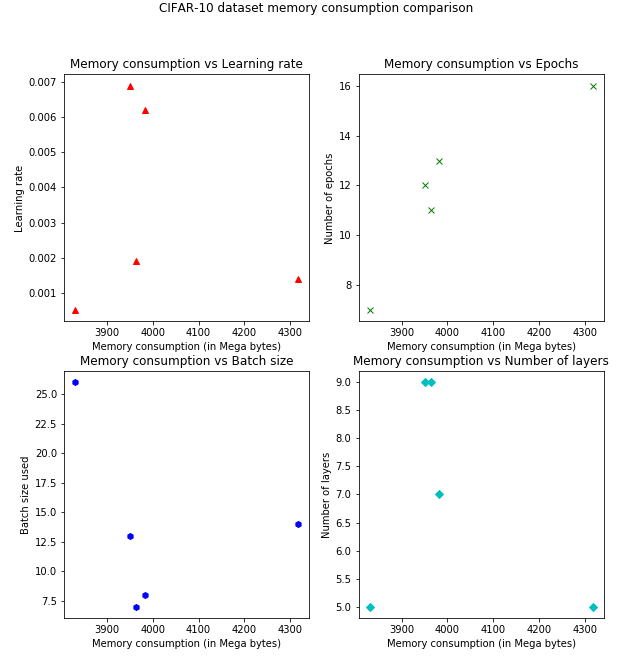}
    \caption{Memory consumption is displayed as graph for comparing neural network with the hyperparameter values for CIFAR-10 dataset.}
    \label{fig:cifar10res}
\end{figure}

The values are used to train the neural network in the back end. The memory consumption values are taken and average is calculated. With the calculated values, the relation between the hyperparameters and the memory consumption can be identified using the graphs in Figure \ref{fig:cifar10res}

This experiment is performed on same setup after some time. The previous experiments for MNIST, Fashion MNIST is performed at a stretch. This clearly indicates that the background memory consumption is higher at this time. In MNIST and Fashion MNIST dataset, the data occupies much smaller space in the images when compared to the CIFAR-10 dataset. This makes the deep learning model to learn the feature easily. Background process is common when all the trails are performed. Based on the experiments performed, higher nodes per layer increases the memory consumption. This is evident from the first and last trial. Calculations done at each epoch depends on the nodes in neural network, depth and rate for convergence depending on learning rate.

Calculating and knowing the memory used can help the developers to analyze the model better and can avoid unnecessary training time if model abruptly utilize more memory. Bigger models can take hours of training time. So knowing the memory utilization in initial phase will be useful for training the model.

\section{LIMITATIONS}

The data extracted related to memory consumption is based on the \textit{psutil} tool. At the same time, we try to maintain that the \textit{psutil} tool only measures the training process of a neural network. There is a possibility that it might also measure other running processes in the system. However, we try to make sure that other applications are not running during the time of the experiments. Hence, the accuracy of gathered memory consumption data during neural network training would be high.

The analysis provided in each section does not provide a concrete pattern in memory consumption of deep learning model. Comparing the hyperparmeters with memory consumption maynot provide valid output although hyperparameters are involved in developing a better accurate model. Also, only three types of standard datasets are used for the experiments. Using a more variety of datasets might bring different insights in understanding the role of the dataset in high memory consumption of deep learning models during its training phase.

\section{RESULTS}

There has been an increase in the use of deep learning models in recent years due to the availability of high computational resources. However, in the view of creating less computational model which aims to optimize the training in difficult environments like within software development, the need to optimize the deep learning models is also increased. In this paper, we tried to look at this aspect and focused on understanding the memory consumption of deep learning models during their training phase. In this regard, we observe the memory utilization while training that model in the real-time. We also performed some experiments to track the memory usage of different neural network models trained on different datasets such as MNIST, Fashion MNIST, and CIFAR-10 with varying hyperparameters. Based on the results, we observed that the learning rate and nodes in hidden layer are the major factors contributing to high memory consumption. However, more investigation needs to be carried out to comprehend the memory consumption in deep learning models.

\section{FUTURE WORK}
In future developments, we will try to find better ways to know the cause of memory utilization in a more detailed manner. If a particular set of nodes causes more memory in a neural network, then there could be measures to remove the nodes for developing better memory consumption models. The proposed approach to track the memory consumption can be extended to other machine learning algorithms and GANs for identifying the cause of using more memory. We propose the perspective of considering memory consumption while developing neural networks. In another perspective, the project can be extended to include energy utilization while developing machine learning algorithms. Energy is an important aspect for training bigger models. So there is a space for larger improvement in this field. Upcoming research to reduce energy consumption is the key motivation for this project. This project paves the way for considering the memory and energy in handling machine learning and deep learning models. Analyzing energy and memory consumption can produce a better understanding of models. This can be extended to reduce the energy and memory consumption of deep learning models in the future.

\bibliography{sample_paper}

\begin{thebibliography}{}

\bibitem[Chen et~al., 2016]{Chen16}
Chen, T., Xu, B., Zhang, C., and Guestrin, C. (2016).
\newblock Training deep nets with sublinear memory cost.
\newblock {\em arXiv preprint arXiv:1604.06174}.

\bibitem[Dayarathna et~al., 2015]{Dayarathna15}
Dayarathna, M., Wen, Y., and Fan, R. (2015).
\newblock Data center energy consumption modeling: A survey.
\newblock {\em IEEE Communications Surveys \& Tutorials}, 18(1):732--794.

\bibitem[Esteva et~al., 2019]{esteva19}
Esteva, A., Robicquet, A., Ramsundar, B., Kuleshov, V., DePristo, M., Chou, K.,
  Cui, C., Corrado, G., Thrun, S., and Dean, J. (2019).
\newblock A guide to deep learning in healthcare.
\newblock {\em Nature medicine}, pages 24--29.

\bibitem[Gao et~al., 2020]{Gao20}
Gao, Y., Liu, Y., Zhang, H., Li, Z., Zhu, Y., Lin, H., and Yang, M. (2020).
\newblock Estimating gpu memory consumption of deep learning models.
\newblock {\em In Proceedings of the 28th ACM Joint Meeting on European
  Software Engineering Conference and Symposium on the Foundations of Software
  Engineering}, pages 1342--1352.

\bibitem[García-Martín et~al., 2019]{garcia2019estimation}
García-Martín, E., Rodrigues, C., Riley, G., and Grahn, H. (2019).
\newblock Estimation of energy consumption in machine learning.
\newblock {\em Journal of Parallel and Distributed Computing}, 134:75--88.

\bibitem[Grigorescu et~al., 2020]{grigorescu20}
Grigorescu, S., Trasnea, B., Cocias, T., and Macesanu, G. (2020).
\newblock A survey of deep learning techniques for autonomous driving.
\newblock {\em Journal of Field Robotics}, pages 362--386.

\bibitem[He et~al., 2016]{Kaiming16}
He, K., Zhang, X., Ren, S., and Sun, J. (2016).
\newblock Deep residual learning for image recognition.
\newblock {\em In Proceedings of the IEEE conference on computer vision and
  pattern recognition}, pages 770--778.

\bibitem[Kim et~al., 2015]{Kim15}
Kim, Y., Park, E., Yoo, S., Choi, T., Yang, L., and Shin, D. (2015).
\newblock Compression of deep convolutional neural networks for fast and low
  power mobile applications.
\newblock {\em arXiv preprint arXiv:1511.06530}.

\bibitem[Krizhevsky et~al., 2012]{Krizhevsky12}
Krizhevsky, A., Sutskever, I., and Hinton, G. (2012).
\newblock {I}magenet classification with deep convolutional neural networks.
\newblock {\em Advances in neural information processing systems}, pages
  1097--1105.

\bibitem[Liu et~al., 2018]{Liu18}
Liu, C., Bellec, G., Vogginger, B., Kappel, D., Partzsch, J., Neumärker, F.,
  Höppner, S., Maass, W., Furber, S., Legenstein, R., and Mayr, C. (2018).
\newblock Memory-efficient deep learning on a spinnaker 2 prototype.
\newblock {\em Frontiers in neuroscience}, 12:840.

\bibitem[Liu et~al., 2019]{Liu19}
Liu, Z., Tang, H., Lin, Y., and Han, S. (2019).
\newblock Point-voxel cnn for efficient 3d deep learning.
\newblock {\em arXiv preprint arXiv:1907.03739}.

\bibitem[Louizos et~al., 2017]{Louizos17}
Louizos, C., Ullrich, K., and Welling, M. (2017).
\newblock Bayesian compression for deep learning.
\newblock {\em arXiv preprint arXiv:1705.08665}.

\bibitem[Nguyen et~al., 2020]{Nguyen20}
Nguyen, D., Hung, N.H.and~Kim, H., and Lee, H. (2020).
\newblock An approximate memory architecture for energy saving in deep learning
  applications.
\newblock {\em IEEE Transactions on Circuits and Systems I: Regular Papers},
  67(5):1588--1601.

\bibitem[Pleiss et~al., 2017]{Pleiss17}
Pleiss, G., Chen, D., Huang, G., Li, T., van~der Maaten, L., and Weinberger, K.
  (2017).
\newblock Memory-efficient implementation of densenets.
\newblock {\em arXiv preprint arXiv:1707.06990}.

\bibitem[Sun et~al., 2020]{Sun20}
Sun, X., Wang, N., Chen, C., Ni, J., Agrawal, A., Cui, X., Venkataramani, S.,
  El~Maghraoui, K., Srinivasan, V., and Gopalakrishnan, K. (2020).
\newblock Ultra-low precision 4-bit training of deep neural networks.
\newblock {\em Advances in Neural Information Processing Systems}.

\bibitem[Tokui et~al., 2019]{Tokui19}
Tokui, S., Okuta, R., Akiba, T., Niitani, Y., Ogawa, T., Saito, S., Suzuki, S.,
  Uenishi, K., Vogel, B., and Yamazaki~Vincent, H. (2019).
\newblock Chainer: A deep learning framework for accelerating the research
  cycle.
\newblock {\em In Proceedings of the 25th ACM SIGKDD International Conference
  on Knowledge Discovery \& Data Mining}, pages 2002--2011.

\bibitem[Vanhoucke et~al., 2011]{Vanhoucke11}
Vanhoucke, V., Senior, A., and Mao, M. (2011).
\newblock Improving the speed of neural networks on cpus.

\bibitem[Zhang et~al., 2019]{Zhang19}
Zhang, J., Yeung, S., Shu, Y., He, B., and Wang, W. (2019).
\newblock Efficient memory management for gpu-based deep learning systems.
\newblock {\em arXiv preprint arXiv:1903.06631}.

\end{thebibliography}

\end{document}